\documentclass[letterpaper, 10 pt, conference]{ieeeconf}  

\IEEEoverridecommandlockouts                              

\usepackage{graphicx}

\title{\LARGE \bf
From Crystallized Adaptivity to Fluid Adaptivity in Deep Reinforcement Learning --- Insights from Biological Systems on Adaptive Flexibility
}

\author{Malte Schilling$^{1}$, Helge Ritter$^{1}$, Frank W. Ohl$^{2}$
\thanks{*This research was supported by the Cluster of Excellence Cognitive Interaction Technology ÔCITECÕ (EXC 277) at Bielefeld University, which is funded by the German Research Foundation (DFG), and the consortium project Intentional Anticipatory Interactive Systems ÔIAISÕ at Magdeburg University and Leibniz Institute for Neurobiology.}
\thanks{$^{1}$Malte Schilling and Helge Ritter are with Center of Excellence `Cognitive Interaction Technology', Bielefeld University, Germany,
        {\tt\small mschilli@techfak.uni-bielefeld.de}}%
\thanks{$^{2}$Frank W. Ohl is with the Department of Systems Physiology of Learning, Leibniz Institute for Neurobiology and with the Institute of Biology, Otto-von-Guericke University, Magdeburg, Germany.}%
}

\begin{document}

\maketitle
\thispagestyle{empty}
\pagestyle{empty}

\begin{abstract}
Recent developments in machine-learning algorithms have led to impressive performance increases in many traditional application scenarios of artificial intelligence research. In the area of deep reinforcement learning, deep learning functional architectures are combined with incremental learning schemes for sequential tasks that include interaction-based, but often delayed feedback. Despite their impressive successes, modern machine-learning approaches, including deep reinforcement learning, still perform weakly when compared to flexibly adaptive biological systems in certain naturally occurring scenarios. Such scenarios include transfers to environments different than the ones in which the training took place or environments that dynamically change, both of which are often mastered by biological systems through a capability that we here term ``fluid adaptivity'' to contrast it from the much slower adaptivity (``crystallized adaptivity'') of the prior learning from which the behavior emerged. In this article, we derive and discuss research strategies, based on analyzes of fluid adaptivity in biological systems and its neuronal modeling, that might aid in equipping future artificially intelligent systems with capabilities of fluid adaptivity more similar to those seen in some biologically intelligent systems. A key component of this research strategy is the dynamization of the problem space itself and the implementation of this dynamization by suitably designed flexibly interacting modules.
\end{abstract}

\section{Introduction}
Adaptive behavior in many animals and humans is still unparalleled by current artificial intelligence (AI) control approaches. While deep learning (DL) or evolutionary algorithm (EA) approaches produce viable solutions in strictly controlled settings, they are still struggling in several characteristic scenarios. A first such scenario is when existing DL and EA approaches face variability arising from complex situations in which numerous degrees of freedom have to be considered \cite{neftci_reinforcement_2019,sutton2019}. A second typical scenario is the occurrence of unpredicted changes in the task or environments that require higher levels of behavioral flexibility and adaptivity --- in natural contexts often on quite fast time scales as compared to the learning dynamics itself. While current AI approaches often excel at finding solutions in narrowly defined problem spaces, and here can outperform biological systems (for example \cite{Silver_2016}), these do typically not transfer to more natural, open and dynamic scenarios; this also typically includes the step from simulation to real physics.
Generally, research on adaptivity and learning in artificial and biological systems has been shown to be mutually fruitful (e.g. \cite{neftci_reinforcement_2019,hassabis_neuroscience-inspired_2017,cully_robots_2015}). In this paper, we want to argue that it might be valuable to consider how the problems solved by current AI approaches and biological systems differ fundamentally. Initially, AI started from static problems that allow for a crisp characterization and showed how learning algorithms can produce solutions by optimizing a cost function. Usually, these cost functions characterize desirable performance criteria for a particular task, such as playing a particular game, such as Go or some of the Atari games, and the learning algorithm is conceptualized as a specific form of adapting to a given task. Typically, this adaptivity unfolds within a huge number of passes through different instances of the given task, and one celebrated breakthrough of AI was the possibility to demonstrate the feasibility of this approach for quite an impressive number of different tasks. However, it has been turned out very difficult to extend this ``crystallized'' adaptivity to also respond quickly to unpredicted variability that occurs on the time scale of a single action: such changes may occur due to unpredicted events during execution of a single task instance, a change to a different task, or a change in the environment, as they are commonplace in most real-world scenarios. Coping with them requires forms of ``fluid'' adaptivity --- captured in the policy --- that can react much quicker than the adaptivity that shapes the skill through the deep learning dynamics (We introduce here the distinction in fluid and crystallized adaptivity that draws on the notion of a distinction in fluid and crystallized intelligence \cite{cattell1971abilities}.).  

Even though there are now approaches that start to take into account selected aspects of fluid adaptivity, from our point of view this notion is even broader and has to deal with further sources of variance. Biological agents in the real word can execute previously learned tasks in different environments. This is not accounted for in approaches dealing with classical games, such as Go or Chess, as in these cases the game is explicitly made invariant to the environment and the rules of the game are embedded into the interaction with the environment. However, physical tasks typically depend strongly on the environment. Dynamically coping with such changes requires correspondingly extended levels of fluid adaptivity. Furthermore, the task itself may change: now fluid adaptivity must become even stretched to cope with an entire range of different tasks. This could lead to a ``cascade'' of fluid adaptivities piled upon each other in a hierarchical fashion. Interactions between such fluid adaptations might give rise to higher-level function, e.g. when predictive components influence each other mutually. 

So far, AI research has primarily focused on crystallized adaptivity and has only recently begun to make some inroads into the layers of this fluid adaptivity hierarchy. Biological systems, in contrast, face very dynamic and constantly changing scenarios that force them from the outset to realize significant fluid adaptivity hierarchies, enabling them to cope with strongly varying contexts and tasks, interact with other animate agents which themselves are adaptive, and to even come up with rapid solutions when the abilities of their body are changed, e.g. due to injuries. While the developed fluid adaptivity works on fast time scales ranging from seconds to fractions of their lifetime, it has been created by a form of crystallized adaptivity that works on much slower time scales, including the very slow time scale of biological evolution. Biological evolution has created brains that complement evolutionary crystallized adaptivity with much faster neural adaptivity that blends learning on the time scales that we see in development and in acquisition of new skills through repetitive training with the fast and reactive fluid adaptivity that we see in reactions to unexpected task or context changes.

Thus, AI and biological intelligent systems foster our understanding of adaptivity from complementary ends:  AI allows us to understand adaptivity by starting out at technically feasible forms of crystallized adaptivity to synthesize solutions to specific tasks, thereby allowing us to work towards increased levels of fluid adaptivity. Biological systems offer us the opportunity to study how entire hierarchies of fluid adaptivity interact in coherent architectures while, however, it is extremely difficult to observe the underlying levels of crystallized adaptivity that have produced these systems.

We want to argue that biological systems indeed provide inspiration on how to overcome the current focus on crystallized adaptivity, i.e. the creation of solutions to given and largely fixed tasks towards finding learning algorithms that can produce an organized hierarchy of fluid adaptivities that support flexible action in a wide range of varying conditions.

Previous work by us and other groups has demonstrated that the described type of flexibility can be studied in several biological systems ranging from insects and rodents (where circuit-level mechanisms can be studied experimentally) to humans (where at least some coarse-grain insights into brain activities are possible by neuroimaging techniques). 

In the following, we turn towards a particularly illustrative example of how current state-of-the-art approaches can successfully solve the complex problem of six-legged robot walking through the introduction of a task-specific behavioral space, but also dramatically fail upon even slight changes in task demands due to lack of fluid adaptivity seen in biological systems. We will then show that biologically inspired approaches that are based on emergent behavior of interacting modules provide a viable solution to such a dynamic problem. In the end, we want to discuss how insights from this and other biologically inspired approaches could help us in designing solutions with similar fluid adaptivity for artificial intelligent systems.

\section{Approaching Deceptive Problems}
Over the last couple of years, deep reinforcement learning (DRL) has been successfully applied in the area of computer games \cite{mnih_human-level_2015}. This area provides a suitable testbed as it allows the learning algorithm to sample a large number of training examples in order to generate good estimates. As more advanced algorithms have been developed, a transfer of such approaches to more real-world problems has begun. Application of DRL in robotics has emerged as an important area \cite{Lillicrap2016ContinuousCW}. In particular, it has become an ongoing research effort to improve algorithms such that they can learn from smaller numbers of training examples.
But such basic DRL approaches suffer from a general shortcoming: DRL is narrowing down towards a single solution. This makes it vulnerable when objective functions show local optima as is often the case. For exploration-based approaches these may give rise to dead-ends from a general strategic viewpoint as they provide no systematic information on how to further improve beyond the currently attained performance level. This means, in order to come closer to a goal state, it might be necessary to first admit a drop in the objective function that only later-on leads to an increased performance. But, unfortunately, this leads back to the problem of how to direct search as we canÕt do an exhaustive search for complex problem spaces.
In the optimization literature such objective functions are called deceptive \cite{Goldberg:1987a} which has been summarized by Lehman and Stanley \cite{lehman_abandoning_2011}: ``a problem is deceptive if lower-order building blocks, when combined, do not lead to a global optimum.'' The general notion of a deceptive problem can be nicely illustrated using the problem of solving a maze. Optimization approaches in general try to string together a sequence of movements (the lower-order building blocks). For reaching a goal location, spatial distance between current and target location has often been used as a simple and effective objective. But in the case of a maze, the single objective of distance towards goal appears too narrow to distinguish between different actions. With a single distance-based objective all locations on a circle around the goal are grouped together and evaluated as having the same value. But this conflation of states and location is a poor aid when trying to find a path through such a maze. The problem is deceptive as moving towards points closer to the goal might actually lead the learning system into a dead-end (for details see \cite{lehman_abandoning_2011}). 

Lehman and Stanley \cite{lehman_abandoning_2011} pointed out that in such an approach stepping stones towards a goal are not rewarded, instead the focus is solely on a single objective which leads the search into local minima. The main problem in deceptive problems is that the chosen objective lumps together quite different states, in the maze example quite different locations, which should be distinguished in order to preserve a structure of the problem space that helps reaching the goal states. Here, one approach from EAs is to consider not a single objective, but to aim initially at what has been called ``stepping stones'' \cite{lehman_abandoning_2011}. This approach is based on a different representation that combines reduction of information from the high-dimensional state space with keeping an intermediate structure that is more effective in directing the system towards an optimal solution than the direct use of the given objective. In the example of finding a path through a maze, stepping stones would introduce an intermediate representation of positional space that conflates places close to each other. . Learning in such an approach is not solely aiming at optimization. Instead, such approaches try to construct an intermediate representation and aim for covering this (therefore not too high dimensional) intermediate representation with stepping stones. This is realized using forms of novelty-based search. In the example of finding a path through a maze, stepping stones would introduce an intermediate representation of positional space that conflates places close to each other. These together would be related towards the objective and when searching for a path towards the goal, decisions for a current state can be based on neighboring stepping stones that are associated towards objective values.
Approaches following such an idea have now been applied in DRL as well. For example, in one hierarchical DRL approach \cite{kulkarni_hierarchical_2016} a similar representation of a positional space in a computer game has been employed as an intermediate level in a hierarchy. Here, the learning agent explored new behaviors for their own sake rather than for immediately solving problems. This approach, reminiscent of ``intrinsic motivation'' in biological systems, later helped the agent to deal with situations when feedback from the environment was either too sparse for a standard reinforcement approach to benefit from it, or when environmental feedback was delayed in time. 

\section{Representations for Stepping Stones --- A Behavior Space}
The above examples have demonstrated in an exemplary fashion that the search for ``stepping stones'' --- either through hierarchical RL or EAs --- can be an efficient strategy for coping with situations in which standard (deep) reinforcement learning approaches fail. Therefore, the question arises if and how such intermediate representations can be found in general and how they would be structured?
Here we start by providing again an example from EAs that has proposed how such a structure might look like. While this allows to cover a range of interesting behavioral problems, we will argue that it still falls short with respect to adaptive behavior and the problem space that animals deal with. The example is taken from six-legged robot walking which is a complex task as it involves a high number of degrees of freedom (at least three per leg and at least 18 overall). Still, even simple animals (with respect to brain size) are capable of very adaptive and unparalleled behavior, climbing through trees, uneven terrain, dealing with leg loss etc. \cite{schilling_walknet_2013}. 
When turning towards motor control approaches, the space of possible behaviors is typically too large for an exhaustive search. A hierarchical approach could help through introducing suitable stepping stones and search for partial solutions to these. This directly leads to the question on what kind of representation a higher level could operate and which would provide a substrate for novelty based search. Cully et al. \cite{cully_robots_2015} demonstrated in a fascinating study how for the example of a six-legged robot (and another example of a robot arm) a suitable problem space representation can be found. The authors argued that animals have an understanding of the whole space of all possible behaviors and that this is shaped by experience in general. Furthermore, in animals this space can be extrapolated towards novel behaviors in novel or changing contexts, e.g. when having to deal with a changed morphology. This behavioral space is evolved in a crystallized adaptivity fashion. In their view, fluid adaptivity is realized as a form of switching between behaviors in the behavioral space.

Cully et al. \cite{cully_robots_2015} introduced a specific, low-dimensional behavioral space that could act as a representational structure for finding stepping stones. In their case, a temporal description of the behavior of all legs was used as a lower (6-) dimensional behavior space (they used duty cycles of the individual legs). Subsequently they trained, using an EA, a behavior performance map. The goal was to cover the low dimensional behavioral space and relate each behavior description towards an optimized behavior, i.e. a high (36-) dimensional action. The optimization was done using an evolutionary approach called multi-dimensional archive of phenotypic elites (MAP-Elites \cite{mouret_illuminating_2015}). MAP-Elites produce a large diversity of high-performing, but qualitatively different solutions. As this approach explores more of the search space concurrently, it also converges towards better overall solutions than many other search algorithms. Such an approach that creates a complex behavior space mapping towards detailed actions was shown to be effective when dealing, for example, with changing morphology. After losing a leg the low dimensional behavior space was used in what they called intelligent trial and error learning: the behavior space was used for variations and different behaviors were tested. The structure of the low dimensional behavior space was used to converge towards a suitable motor control program.

\section{Dynamic Problem Spaces --- Fluid Adaptivity in Biological Systems}
The approach by Cully et al. \cite{cully_robots_2015}  has shown to work efficiently for changes of morphology. And this provides an important first step: in real world scenarios, we have to deal with changes of an environmental setting and in particular the environmentÕs response to the actions of an agent might change. This is difficult for all control approaches and in the case of a formalization of reinforcement learning as one holistic agent observing interactions with the environment taken as one singular unit this becomes problematic.

\begin{figure}[tbp]

\includegraphics[scale=0.3]{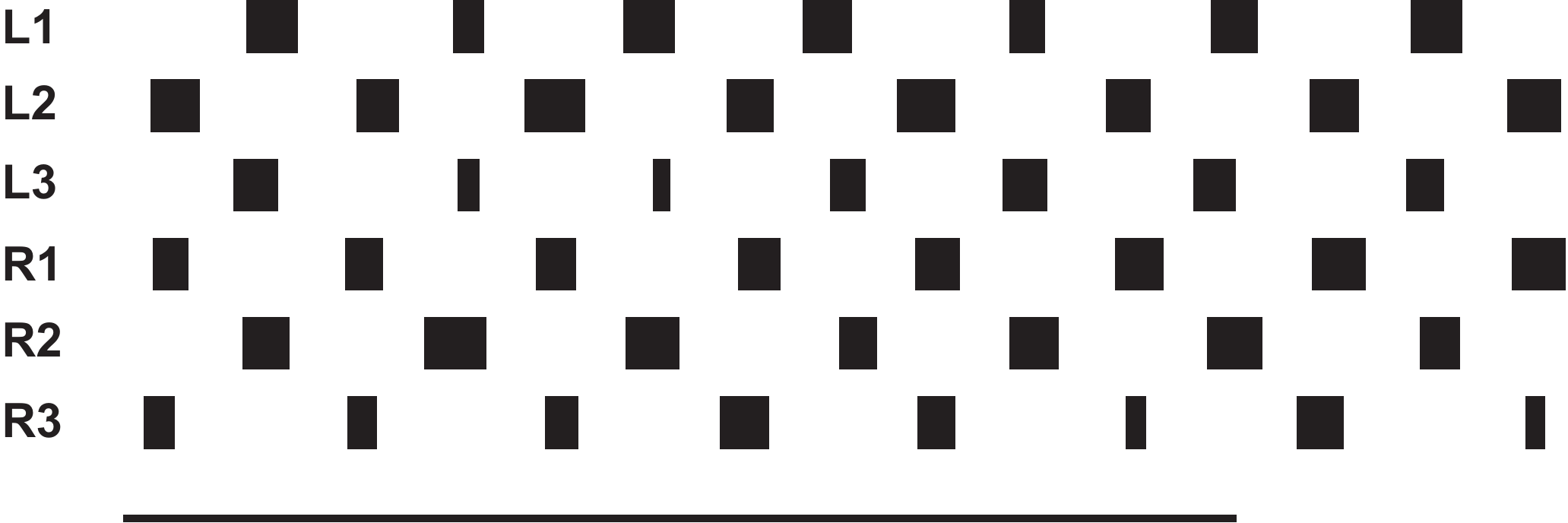}

\vspace{12pt}

\includegraphics[scale=0.3]{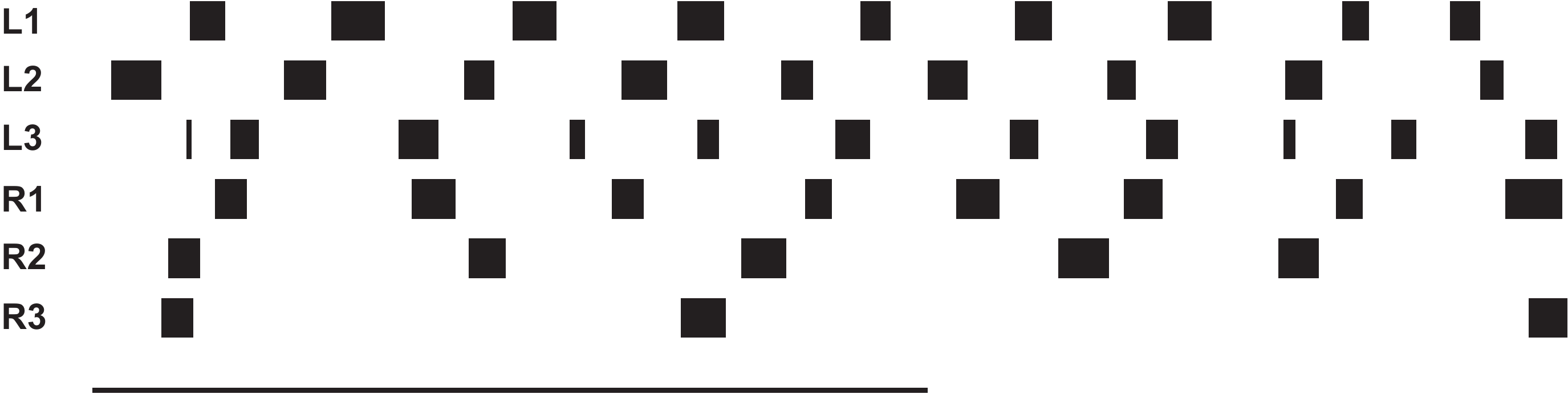}

\caption{Top -- Footfall pattern for tripod gait, black bars indicate swing movement of the respective leg (L1: front left leg, L2: middle left leg, L3: hind left leg, R1: front right leg, R2: middle right leg, R3: hind right leg), abscissa is simulation time. The lower bars indicate 500 iterations corresponding to 5s real time. Bottom -- footfall pattern for curve walking (turn towards the right). For more information see \cite{schilling_hexapod_2013}. }
\label{fig_curve}
\end{figure}

It should be noted that the example of morphology change considers only one singular change. In contrast, many real world problems are characterized by ongoing change and the fact that the whole problem space is dynamically changing all the time. Nevertheless, this is the kind of problem space that biological system can often excel at due to their highly developed fluid adaptivity.
We think that the example of six legged walking can help to demonstrate this point and want to explain in which ways coping strategies for highly dynamic environments in insects, well-investigated hexapod walkers, differ from the above considered types of a learned behavior performance map.
Central to the approach in \cite{cully_robots_2015} is how the behavioral space factorizes the state space. Describing temporal coordination appears as a meaningful representation. Such temporal patterns are traditionally used, in order, to describe how animals move (often plotted over time as footfall patterns, see Figure). While we agree that these patterns provide a highly suitable observational space we want to argue this space might not be used for motor control. Central to our argument is that, especially in six-legged walking animals, fixed gait patterns are typically not observed, but gaits showing high levels of variability. Fixed patterns can mostly be observed during running or walking with a high speed and, in particular, on flat terrains \cite{schilling_walknet_2013}. In uneven terrains or while climbing through twigs the environment demands for spatial adaptation of footholds which in turn requires permanent temporal and spatial adjustments. Even in slower walking, when gaits emerge, these are more typically characterized by a spectrum of temporal locomotion patterns. Consequently, in such scenarios, walking cannot be straightforwardly be described by phase relations between different legs as those are constantly changing and adapted. A description by fixed patterns would be even harder when just duty cycles were used (cf. \cite{cully_robots_2015}). Generally, fixed relations are widely used because of their simplicity and the fact that they provide good walking results on flat terrains and when maintaining a certain velocity. But fixed gaits run into problems during acceleration or deceleration, i.e. in conditions when something is (rapidly) changing. The approach by Cully et al. \cite{cully_robots_2015} appears well equipped to adjust after a change, but this type of representation appears not well suited when it comes to moderating constant change. As one example, consider how temporal coordination (and duty cycles as well) drastically changes in natural behaviors all the time (see Fig. \ref{fig_curve}): when negotiating a curve the temporal coordination between legs and their respective duty cycles are under constant change (see Fig. \ref{fig_curve} on the bottom, e.g. the right hind leg is barely moving and provides an anchor point for the turn towards the right; in even tighter turns it may stand still and wonÕt produce any stepping during the curve).

From our point of view, the approach by Cully et al. \cite{cully_robots_2015} nicely points out that a hierarchical organization and structuring of the behavioral space helps in finding solutions for complex motor control problems. But, on the other hand, we think that the idea of a static, stationary representation might be too restricting, to address the whole range of behavioral adaptivity. It appears that the footfall pattern emerges as a form of an observable behavioral code, but it seems questionable as a substrate for control. So, while such a lower-dimensional space appears helpful to make dynamic problems (ranging from small changes in the environment towards large, constantly changing environments) accessible, we would argue that a static given representation appears too narrow and too much tuned towards a specific kind of behavior.
Instead, we would like to suggest to consider a slightly different take on such a behavioral space which does not treat the behavioral space as a specific type of representation, but rather as an observation of invariances that stem from the underlying systems. In six-legged walking such an approach has shown to be very promising with respect to adaptivity. Our Walknet approach consists of six decentralized control structures, one for each leg which makes decisions on which action (swing or stance movement) to execute for that specific leg \cite{schilling_walknet_2013}. The overall behavior of the system emerges from local interactions between these controllers. For an outside observer, such a system produces a form of stable temporal patterns (as shown in Fig. \ref{fig_curve}, top). Importantly, these are highly adaptive. When the walking system is externally disturbed (either through a spatial disturbance as when walking over uneven terrain or a temporal disturbance when a leg is not finding a foothold or is actively displaced) the observed pattern might look very different, but overall the interacting modules together compensate for the disturbance and overall the system quickly settles back towards a well-organized walking pattern and stable walking without tumbling. In the following part, we will argue that it might be more appropriate to understand such motor control architectures as active, interacting control structures, instead of trying to find specific representations that mediate between different modules. While such concurrent control circuits and their dynamic interactions may be more difficult to analyze and understand, they might offer a better starting point for approaching adaptivity, especially in more complex hierarchically organized systems. Instead of enforcing a particular representation scheme, the interaction between the different (simple) dynamical systems span a behavioral space that provides a rich substrate for adaptive behavior.

\section{Fluid Adaptivity as a Challenge for DRL}
We have considered finding stepping stones in a behavioral space in an EA approach. Again, these ideas are finding now their way into DRL and it shows that they can be successfully leveraged to address more complex problems. Maintaining diverse and multiple solutions concomitantly active appears beneficial with respect to changes in the environment. This has already been successfully incorporated into reinforcement learning. Hierarchical approaches are applied in meta-learning and transfer learning settings. These aim for generalization between different, but related tasks and try to reuse existing learned structure between tasks. 
Importantly, the structure of a hierarchy allows to not only converge towards one single solution, but to divide the problem and maintain different solutions that can be activated when appropriate (similar to population-based approaches \cite{DBLP:journals/corr/abs-1807-01281} in which a whole population of different behavioral strategies is evolved). But still, these approaches typically require quite a lot of training and are therefore mostly applied in dynamic simulations of the robotic agent. The problem of simply dealing with a variable real-world setting in a sample-efficient way still remains a challenge. Trying to capture the variability becomes increasingly difficult through simple exploration of inherently variable responses from the environment. This becomes even more of a problem when there are other active agents present which might learn to react towards our actions or simply vary how to act and what goals to follow. 

In the following, we want to explain why considering a permanently dynamically changing environment (its response characteristic) is problematic for the employed conceptualization as a simple Markov Decision Problem (or a semi-MDP in the case of a hierarchical model) and why therefore following such a route might not scale as well as currently hoped for.

One important insight for DRL approaches was the idea of how to use past experience to train a value function using TD-learning. Such value functions are supposed to provide an estimate of a next state (or a chosen action), but this value is always necessarily conditioned on an assumption how one would proceed from there on. Whenever one starts to change the estimator for a value function which is exploited for subsequent action selection, this might interfere with this estimated assumption of what would follow. This problem of off-policy learning has shown to be a serious problem for DRL and often leads to instabilities that cause complete breakdowns of such learning approaches. In particular, overestimation of values has shown to draw learned control approaches towards such over-estimated actions and often learning never recovers from such bad selections. Different solutions have been proposed, for example using separate and more conservative estimates for action selection \cite{hasselt_deep_2016}. One alternative would be to always require gathering new experience which, unfortunately, would be very costly and unfeasible for many problems. Even worse, and here we turn towards the distinction of different types of problems: in many real world settings, agents not only have to deal with a novel, given situation. But part of it are other agents that are learning themselves. This means that in these cases the response of the environment might change by itself and independently of the changes in the learning agent itself. Therefore, the agentÕs current policy is already rendered in a way off-policy as he is trying to optimize towards responses of the environment which have changed themselves. From the experience in DRL it is known that this destabilizes learning an estimator and that this makes it difficult to handle or even intractable. Current approaches trying to alleviate this problem, for example, similar to the DDQN approach, a form of correction has been applied in continuous learning settings \cite{fujimoto_addressing_2018}. This allows to push the border further and deal with some more variability of behavior. But as soon as the environment is allowed to change, such approaches run into trouble. 

The fundamental problem of a dynamically changing environment is not addressed at its core. Going back to the simple maze problem: consider a change of the maze, like adding a wall that closes the shortest path in a maze or adding a novel high reward at a different location. For reinforcement learning, all the paths that cross through that passage would be destroyed and all values that depend on paths (in q-learning through the max operation) through this passage would be misleading. Closing the passage ``revivesÕÕ the relevance of experiences from alternative paths whose values  --- in the presence of the passage --- were dominated by the better values of paths through the passage. However, standard formulations of RL have no memory for such suboptimal paths; their contributions thus become overwritten as soon as a better alternative is detected. They thus cannot become ``revivedÕÕ when the maze changes to a condition where they would be again relevant. In contrast, biological agents can handle dynamic environments that in themselves are constantly acting and reacting. For example, in Huh et al. \cite{huh_model-based_2009} rats were trained in a free-choice task in a maze with multiple targets that probabilistically provided a reward. In one condition, the probability for a reward increased for the unchosen target. The animals learned this change of probabilities. Importantly, this could not be accounted for by a simple reinforcement learning approach that learns a sequence of decisions. But it was found better accounted for using a model-based reinforcement learning approach.
Similar results have been obtained in research on reversal learning, where a learning agent first learns a set of appropriate behaviors for a given set of situations and then, suddenly, has to show different associations of learned behaviors to the set of situations, in order to reach behavioral goals. While standard RL agents have to be re-trained entirely in such reversal scenarios, biological systems are able to remap hierarchically organized structures of representations apparently created during initial training, and solve subsequent reversal problems much more effectively \cite{jarvers_reversal_2016}. There is evidence that biological brains solve such problems by an efficiently structured communication between separable brain modules that implement RL strategies and those that implement knowledge representation \cite{schulz2016}.
 These neuronal strategies in biological systems are reminiscent of model-based RL approaches. The general idea of model-based reinforcement learning is to ``decompose the RL problem into two subproblems: learning a model of the environment, and then planning with this model'' \cite{silver_predictron:_2016}. Model-based reinforcement learning has become a prominent approach. We will not consider this approach in detail here, but want to remark, as a side note, that the view of dividing the problem into two subproblems will suffer from similar shortcomings we discussed before. When trying to learn the interaction with the environment as one holistic model, this will require a long time as there are many dimensions to this problem. Such a model would probably try to capture too much variability.

Alternatively, we propose to follow a more modular approach in which models are learned in the environment and in which the focus is on the compositional semantics of how these modules can interact. Note that such an approach takes a different perspective on reinforcement learning that is breaking up the standard and simple notion of an agent interacting with one holistic environment. We next turn to a better suited factorization of multiple active modules and their compositional semantics.

\section{Towards Dynamic Problem Spaces}
As the last aspect, we want to briefly discuss a possibility how such a slightly changed reinforcement learning perspective might look like, again exemplifying our argument with the example of hexapod walking. The EA approach by Cully et al. \cite{cully_robots_2015} introduced a representation that was well suited for the problem of changing morphology. We argued that for negotiating a curve this representation appears not optimally suited and another form or representation would be required. As there are probably many and quite different problems six-legged walking agents will face, it would be naive to require a specific representation for each dimension of adaptive behavior. Instead a better approach might be to understand how the behavior emerges from interactions of different low-level control modules. Luckily, we have the well-grounded model of insect behavior in which behavior emerges from decentralized and concurrent control structure. Basically, each leg is assumed to be controlled by an independent motor controller and there are only a couple of local coordination influences. Interaction of six such local controllers can explain a wide variety of behaviors, including dealing with change of morphology and negotiating curves \cite{schilling_hexapod_2007,schilling_walknet_2013}. This approach differs fundamentally from the EA approach. In the EA approach a fixed behavior space was defined and behaviors were optimized over this lower dimensional space --- behaviors were evolved as a form of crystallized adaptivity. But a so-defined behavioral space appears limited in the way how it can deal with the overall adaptivity of insect behavior \cite{schilling_walknet_2013,dickinson_how_2000} which is an example of fluid adaptivity. Insect behavior is adaptive on the very short timescale and deals with (to a large degree unpredictable) variability of the environment in a flexible way. Temporal coordination appears to emerge from concurrent local computations and sensory interactions with the environment. In the decentralized model based on biological data from insects, we also can, at any given time, map behavior towards the behavior space for descriptive purposes but we don't assume that is used for motor control. Instead, we have access to the underlying processes that generate this behavioral space (without requiring more dimensions). 
This control approach has been used up to now as a hand-crafted control system. Extending this towards reinforcement learning appears as another promising direction: learning novel behaviors would be a next step. But, in this case a hierarchy would not be realized on a representational level introducing a fixed representation for a behavioral space, but on a control level: there are underlying concurrent processes which can interact --- the overall problem would be modularized. We would still be able to observe representations similar to those used by Cully et al. \cite{cully_robots_2015}, but not as a level in a control hierarchy. Instead, these would emerge from the hierarchical control system and, when turning towards a problem like curve walking, we would be able to observe quite different and better suited emerging representations. This would require a reformulation of the hexapod approach as was done in \cite{schilling2018ddrl}. The reformulation is based on (partially) independent concurrent control system and the whole problem becomes similar to a multi-agent scenario. But introducing such a factorization appears promising and should as well be able for cooperative multi-agent settings.

\section{Discussion and Conclusion }
In this paper we presented the examples of six-legged walking and reversal learning as two illustrative scenarios in which the solutions found by biological agents still outperform our best models based on DRL. To analyze this situation we found the terms crystallized adaptivity and fluid adaptivity helpful to describe, respectively, the optimization of agent behavior with respect to a well-defined problem, or the preparation of the agent enabling it to cope with variations to the previously learned strategies. While biological and artificial systems might show high comparability with respect to crystallized adaptivity (cf. \cite{cully_robots_2015}), established learning algorithms used in engineering usually fail to show the level of fluid adaptivity that characterizes biological agents (e.g. \cite{schilling_walknet_2013,jarvers_reversal_2016}). Previous studies have already established that significant performance improvements in evolutionary or learning algorithms can be achieved by considering (via discovery or construction) a new intermediate level of representation in addition to the original problem space. But typically, even this measure, failed to equip the agent with a sufficient level of fluid adaptivity to mimic the flexibly adaptive behavior seen in biological systems. Our research points into the direction that a different approach might be fruitful, namely that flexibly adaptive solutions can be found by proper factorization of this intermediate representation. The factorization can be generally implemented by a set of parallel modules, and potentially a hierarchy of such parallel sets, that are independently trained by interaction with the environment. This interaction can in fact be decoupled from the usual reinforcement learning updating cycle. As evidenced by our examples it is the suitably organized interaction semantics between these modules that then provides the ability of the agent to flexibly respond to new environments of fast changes of conditions within a given environment both of which were not encountered during training.

It might also be interesting to put our results into context with other current developments in AI research: Recently, DeepMind used the competitive computer game StarCraft successfully as a testbed for training DRL approaches \cite{vinyals2019} and in their case they reframed the problem using a game-theoretic approach. First, and this is a lesson we covered in the article as well, it showed beneficial to maintain a population of multiple solutions that might fit at different times (\cite{balduzzi_open-ended_2019} and as pointed out in \cite{arulkumaran_alphastar:_2019} which provide a complementary EA perspective on AlphaStar). But, importantly, they employed a different kind of representation for their novelty based search. In order to formalize the diversity in the population, Dadashi et al. \cite{dadashi_value_2019} introduced the gamescape which (as they showed) is a polytope in the space of projections of policies towards value functions. Interestingly, there is structure to this space that they (apparently) exploited. This offers a promising view as the interaction with the environment is formalized as dynamic --- the environment is able to respond in different ways and this is learned. Such a modularization of the problem appears as a promising direction. 
More generally, progress in the attempt to equip artificial systems with similar capabilities of fluid adaptivity as seen in some biological systems, might depend on the results of deeper research on how this modularization of the problem space and the interaction semantics of the resulting modules can be generally developed, beyond specific application scenarios. In this context, it appears important to further search for factorizations of the overall problem space of interactions of the environment with an agent. This has to consider that both, agent and environment, are dynamic. But here it might be beneficial to take into account some insight into
the structure of biological systems. 
For the control of adaptive behavior, one important structural consideration could be a modular decomposition of control systems and dynamic environments.


\bibliographystyle{IEEEtran}
\bibliography{References}

\end{document}